\documentclass[oneside,twocolumn,ngerman]{article}

\usepackage{amsmath,amssymb}
\usepackage{graphicx}
\usepackage{tikz,pgfplots}
\usepackage{caption}
\usepackage[shortlabels]{enumitem}
\usepackage[font=small]{subcaption}
\usepackage{float}
\usepackage{hyperref}
\usepackage{booktabs}
\usepackage{cuted}
\usepackage{multirow}
\usepackage{selinput}
\usepackage[T1]{fontenc}

\SelectInputMappings{germandbls={ß}}

\newcommand{\norm}[1]{\left\lVert#1\right\rVert}

\title{\vspace{-2.5cm}{\bf Computational characteristics of feedforward neural networks for solving a stiff differential equation}}
\author{Toni Schneidereit and Michael Breuß \\ 
\\[1ex]
\small Applied Mathematics Group \\ 
\small Brandenburg University of Technology Cottbus-Senftenberg \\
\small Platz der Deutschen Einheit 1, 03046 Cottbus, Germany \\
\small \{Toni.Schneidereit,breuss\}@b-tu.de
}

\date{\small \today}

\begin{document}

\twocolumn[
\begin{@twocolumnfalse}
\maketitle
\begin{abstract}
Feedforward neural networks offer a possible approach for solving differential equations.
However, the reliability and accuracy of the approximation still represent delicate
issues that are not fully resolved in the current literature.
Computational approaches are in general highly dependent on a variety
of computational parameters as well as on the choice of optimisation methods,
a point that has to be seen together with the structure of the cost function.
The intention of this paper is to make a step towards resolving these open issues.
To this end we study here the solution of a simple but fundamental stiff ordinary differential equation
modelling a damped system. We consider two computational approaches for solving differential equations
by neural forms. These are the classic but still actual method of trial solutions defining the cost function,
and a recent direct construction of the cost function related to the trial solution method.
Let us note that the settings we study can easily be applied more generally, including solution of partial
differential equations.
By a very detailed computational study we show that it is possible to identify preferable choices
to be made for parameters and methods. We also illuminate some interesting effects that are observable in
the neural network simulations.
Overall we extend the current literature in the field by showing what can be done
in order to obtain useful and accurate results by the neural network approach.
By doing this we illustrate the importance of a careful choice of the computational setup.
\end{abstract}
\begin{center}
\vspace{0.875cm}
\end{center}
\end{@twocolumnfalse}]

\section{Introduction}

Differential equations occur in many fields of science and engineering and represent a useful description of many physical phenomena.
They are usually formulated as initial or boundary value problems, where conditions at the beginning of a process
or at boundary points are given to obtain one specific solution. It is useful to approximate differential equations
by numerical methods \cite{Hanke2009comp},\cite{Antia2012book} like finite difference methods, and also neural networks have been applied to this end, see e.g., \cite{Kumar2012Appl,Parisi2003unsuperv,Dissanayake1994approx}.

The variety of possible neural network architectures is immense \cite{Leijnen2020zoo}. Already in classic works
in the field, feedforward neural networks have proven to be useful for solving differential equations
\cite{Maede1994lin,Maede1994nonlin}. 
Within this framework of feedforward neural networks (from now on denoted here simply as neural networks),
two particular approaches have been investigated in the literature within the last decades that appear to be
very promising.

The trial solution (TS) method has been proposed for the approximation of a
given differential equation, which we abbreviate here as TSM \cite{lagaris1998artificial}. The TS, also called neural form
in \cite{lagaris1998artificial,Lagari2020Neural}, has to contain the neural network output and has to satisfy given initial or boundary conditions by construction.
Under the latter conditions there are multiple different possible forms of the TS for the approximation of a differential equation.
Recently a systematic construction approach for the TS has been proposed \cite{Lagari2020Neural}. However, as indicated in the latter work, the TS construction
may become difficult to realise for complex problems. We will follow here the original approach proposed in \cite{lagaris1998artificial}.
Let us note that the same TS structure is used for example in the recent Legendre neural network \cite{mall2016legendre}.
It appears evident that our investigation may also be useful in the context of such extensions.

Recently published in 2019, an approach has been proposed to avoid finding a TS \cite{piscopo2019solving} as this may be an intricate
ingredient of the TSM. Because the corresponding
method is motivated and technically related to the TSM, we call it here modified trial solution method (mTSM).
Instead of building the cost function by use of the TS that meets conditions imposed on a differential equation,
the approximating solution function is set in \cite{piscopo2019solving} to be the neural network output directly.
The latter does not satisfy given initial or boundary conditions by construction as in TSM, but these are added
as additional terms in the cost function.

In the mentioned works, both TSM and mTSM have proven to be capable of solving ordinary (ODEs) and partial differential equations (PDEs)
as well as systems of ODEs and PDEs. This has been demonstrated for several examples and even complex simulations,
showing the potential of the methods to obtain high-quality results. This has motivated us to consider in higher detail some of the computational
issues that arise in the application of these methods in a first study \cite{Schneidereit2020ODEANN}.
Let us mention here also a recent complementary work where the activation functions are subject of a
computational study \cite{Famelis2020study}.

Despite these promising developments, there are still many open questions related to both TSM and mTSM. First of all, in the
original work \cite{piscopo2019solving} an emphasis was layed on the new construction principle and the application of the proposed method
in a cosmological context. However, one may wonder about the direct comparison of TSM and mTSM in terms of quality
of results as well as in the related computational aspects. Let us stress in this context, that the original works
\cite{lagaris1998artificial,Lagari2020Neural,piscopo2019solving} mainly describe the network architecture and elaborate on the TS and
mTSM construction, but they do not contain more details of the computational characteristics of the methods. Yet it turns out that it
is not trivial to define a computational framework that gives competitive results.
\\ \newline
{\bf Our contribution.}
In this work we build upon our first parameter study in \cite{Schneidereit2020ODEANN} and extend the investigations.
Following the basic line of the first work, we study here the variance between the exact solution of an ordinary differential equation and the
approximations provided by TSM and mTSM. As one apparent difference to the proceeding in our previous conference paper, we extend here
the investigation w.r.t.\ the number of training points, and we give many more details in the evaluation. We also give here additional and
as it turns out meaningful experiments concerned with the roles of neural network weight initialisation, number of hidden layers and number of hidden layer neurons.
We perform several experiments on the variety of parameters related to the differential equation, neural network and optimisation methods.

Let us stress that the amount of parameters for the differential equation, neural network and optimisation is numerous. Our contribution in
the main part of this paper is a study of the variation on (\textit{i}) Weight initialisation methods, (\textit{ii}) Number of hidden layer neurons,
(\textit{iii}) Number of hidden layers, (\textit{iv}) Number of training epochs, (\textit{v}) Stiffness parameter and domain size,
(\textit{vi}) Optimisation methods, and especially their mutual dependence. Let us note that it has turned out to be a nontrivial task to set up a meaningful proceeding that gives an account of the latter aspect. We consider the evaluation presented here as a number of carefully chosen experiments that are in many respects related to each other.

For investigating the computational characteristics, we consider a simple while important stiff ODE model equation \cite{Dahlquist1978stiff}
with a damping behaviour for studying the stability and reliability of both methods. We also present here as another contribution
a detailed study of the influence of the stiffness parameter contained in the ODE. Let us note that a similar solution behaviour
is to be expected when resolving for instance parabolic PDEs.

\section{Neural network architecture}

\begin{figure}[!h]
\centering
\includegraphics{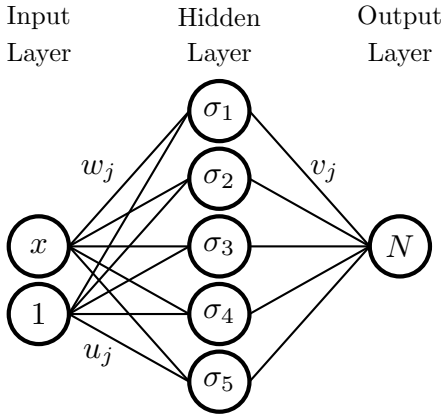}
\caption{Neural network architecture as usually in many of our experiments featuring one input layer, one hidden layer with sigmoid activation functions and one output layer \label{figANN}}
\end{figure}

Neural networks are usually pictured as neurons (circles) and connecting weights (lines). Fig.\ \ref{figANN} shows the standard neural network architecture for our experiments. It consists of three layers and features one input layer neuron for $x\in D\subset\mathbb{R}$ (where $D$ denotes the domain) with one bias neuron which can be considered as an offset, five hidden layer neurons and one linear output layer neuron. In experiments on the number of hidden layers and the number of hidden layer neurons, the architecture is extended. Each
neuron is connected with every single neuron in the next layer by the weights $w_j$, $u_j$ and $v_j$, $j=1,\ldots,5$, which are stored in the weight vector $\vec{p}$. The input layer passes the domain data $x$, weighted by $w_j$ and $u_j$ to the hidden layer for processing. The processed data is then, now weighted by $v_j$, sent to the output layer in order to generate the neural network output $N(x,\vec{p})$. That means in detail, the hidden layer receives the weighted sum $z_j=w_jx+u_j$ as input and processes this data by the sigmoid activation function $\sigma_j=\sigma(z_j)=1/(1+e^{-z_j})$. Since the output layer consists of a linear neuron, the neural network output is generated by the linear combination
\begin{equation}
N(x,\vec{p})=\sum_{j=1}^{5}v_j\sigma(z_j). \nonumber 
\end{equation}
The sigmoid activation function is a continuous and arbitrarily often differentiable function with values between 0 and 1.

Let us note, that in order to solve differential equations of order $n$ with neural networks, it is important to choose an activation function, which is at least $(n+1)$ times continuously differentiable, since the later shown solution approaches require the $n$-th activation function derivative and the optimisation methods require another differentiation.

The universal approximation theorem \cite{Cybenko1989uat} states, that one hidden layer with a finite number of sigmoidal activation functions is able to approximate every continuous function on a subset of $\mathbb{R}$. 

In general, it is common to initialise the weights with small random values \cite{Nguyen1990weights}, therefore the first computation of $N(x,\vec{p})$ returns a random value. This value is used to compute the \textit{cost} or \textit{loss} function $E[\vec{p}]$ which is then subject to optimisation.
With the first random output of $N(x,\vec{p})$, the optimisation will return different weight updates
when starting several computations with exactly the same computational parameters (but random weight initialisation).

Another option is to choose the initialisation to be constant when starting several computations.
That is, $N(x,\vec{p})$ first returns always the same value, and therefore with
the weight updates to be constant as well, computations with same parameter setting return equal results. \par

For supervised learning, where both input data $x_i$ (representing the discrete domain or grid) and correct output data $d_i,~i=1,\ldots,n$, are known, the cost function may be chosen as the squared $l_2$-norm    
\begin{equation}
E[\vec{p}]=\frac{1}{2}\big\lVert N(x_i,\vec{p})-d_i\big\rVert_2^2,\nonumber 
\end{equation}
while in case of unsupervised learning, where no correct output data is known, the cost function is part of the modelling process. Our approach follows the latter track.

\section{Solution approaches}

In this section, we will describe the trial solution construction for TSM and mTSM more in detail, as well as the approaches on how to make use of neural networks in order to solve ordinary differential equations (ODEs) in form of
\begin{equation}
G\left(x,u(x),\frac{d}{dx}u(x)\right)=0,~~x\in D\subset\mathbb{R},
\label{GenForm}
\end{equation}
with given initial or boundary conditions. In Eq.\ \eqref{GenForm}, $u(x)$ denotes the exact solution function with $x$ as independent variable. Although $G$ denotes a first order ODE, let us note again that it is also possible to solve higher order ordinary or partial differential equations (PDEs), as well as
systems of ODEs or PDEs, cf.\ \cite{lagaris1998artificial,piscopo2019solving}. 

\subsection{Trial solution method (TSM)}

Let us now recall the approach from \cite{lagaris1998artificial}. In order to satisfy initial or boundary conditions, the TS is constructed to satisfy these conditions and is therefore written as a sum of two terms
\begin{equation}
u_t(x,\vec{p})=A(x)+B(x)N(x,\vec{p}).
\label{TrialSol}
\end{equation}  
In Eq.\ \eqref{TrialSol}, $A(x)$ is supposed to satisfy the initial or boundary conditions at the initial or boundary points, while $B(x)$ is constructed to become zero at these points to eliminate the impact of $N(x,\vec{p})$ there. That is, the TS may be defined in many possible forms for one differential equation, satisfying the mentioned conditions. Especially the choice of $B(x)$ determines the impact of $N(x,\vec{p})$ over the domain. Now, the TS transforms Eq.\ \eqref{GenForm} into
\begin{equation}
G\left(x,u_t(x,\vec{p}),\frac{\partial}{\partial x}u_t(x,\vec{p})\right)=0,
\label{GenForm2}
\end{equation}   
so that the partial derivative of the trial solution with respect to input $x$ which we have to consider is
\begin{equation}
\frac{\partial }{\partial x}u_t(x,\vec{p})=A'(x)+B'(x)N(x,\vec{p})+B(x)\frac{\partial }{\partial x}N(x,\vec{p}) \nonumber
\end{equation}
with 
\begin{equation}
\frac{\partial }{\partial x}N(x,\vec{p})=\sum_{j=1}^{5}v_jw_j\sigma'(z_j). \nonumber
\end{equation}
In order to generate training points for the neural network, we discretise the domain $D$ by a uniform grid
with $n$ grid points $x_i$. Over this discrete domain, Eq.\ \eqref{GenForm2} is now solved by an unconstrained
optimisation problem using the cost function
\begin{equation}
E[\vec{p}]=\frac{1}{2}\norm{G\left(x_i,u_t(x_i,\vec{p}),\frac{\partial}{\partial x}u_t(x_i,\vec{p})\right)}_2^2. \nonumber
\end{equation}

\subsection{Modified trial solution method (mTSM)}

This method, proposed in \cite{piscopo2019solving}, introduces
\begin{equation}
u_t(x,\vec{p})=N(x,\vec{p}), \nonumber
\end{equation}
as a TS directly for all differential equations. Therefore $u_t$ does not satisfy initial or boundary conditions by construction
as in \eqref{TrialSol}, they rather appear in the cost function as additional terms
\begin{align}\nonumber
E[\vec{p}]&=\frac{1}{2}\norm{G\left(x_i,u_t(x_i,\vec{p}),\frac{\partial}{\partial x}u_t(x_i,\vec{p})\right)}_2^2 \\ \nonumber
&+\frac{1}{2}\big\lVert u_t(x_m,\vec{p})-K(x_m)\big\rVert_2^2,\nonumber
\end{align}
where $K(x_m),~m=1,\ldots,l$, denote the initial or boundary conditions.

The modelled cost function is now subject to optimisation with respect to the adjustable neural network weights $\vec{p}$.
 
\section{Optimisation}

For cost function minimisation we use first order methods, based on gradient descent. A commonly employed, simple optimisation technique is
backpropagation, which uses the cost function gradient with respect to the neural network weights to determine their influence on $N(x,\vec{p})$
and to update them. It is well-known that backpropagation enables to find a local minimum in the weight space.
The training is usually done several times with
all training points. After one complete iteration through all input data, one epoch of training is done and for efficient training
(finding a minimum in the weight space), several epochs of training are performed.
For the \textit{k}-th epoch, backpropagation with momentum update rule \cite{Phansalkar1994momentum} reads as   
\begin{equation}
\vec{p}(k+1)=\vec{p}(k)\underbrace{-\alpha\frac{\partial E[\vec{p}(k)]}{\partial \vec{p}(k)}+\beta \Delta\vec{p}(k-1)}_{\Delta\vec{p}(k)}.
\label{updateP}
\end{equation}
Since only the neural network output $N(x,\vec{p})$ and the derivative w.r.t. $x$ depend on $\vec{p}$ and as both expressions are given,
the corresponding derivatives used in the gradient of the cost function are computed as
\begin{align}
\nonumber
&\frac{\partial }{\partial w_j}N(x,\vec{p})=v_jx\sigma'(z_j) \\ \nonumber
&\frac{\partial }{\partial u_j}N(x,\vec{p})=v_j\sigma'(z_j) \\ \nonumber
&\frac{\partial }{\partial v_j}N(x,\vec{p})=\sigma(z_j) \nonumber
\end{align}
and
\begin{align}
\nonumber
&\frac{\partial }{\partial w_j}\left(\frac{\partial }{\partial x}N(x,\vec{p})\right) =v_j\sigma'(z_j)+v_jw_jx\sigma''(z_j) \\ \nonumber
&\frac{\partial }{\partial u_j}\left(\frac{\partial }{\partial x}N(x,\vec{p})\right) =v_jw_j\sigma''(z_j) \\ \nonumber
&\frac{\partial }{\partial v_j}\left(\frac{\partial }{\partial x}N(x,\vec{p})\right)=w_j\sigma'(z_j).  \nonumber
\end{align}
The momentum term in Eq.\ \eqref{updateP}, with momentum parameter $\beta$, uses impact from last epoch to reduce the chance of getting
stuck too early during training in a local minimum or at a saddle point.

The learning rate $\alpha$ in general, is a scaling factor for the
gradient and has major influence on the update. A very basic approach is to choose $\alpha$ as a constant learning rate (cBP).
In order to prevent the optimiser from oscillating around a minimum one may employ a variable learning rate (vBP) as an alternative.
Different approaches for learning rate control exist \cite{Kaneda2015stepsize}, we opt to employ the linear decreasing model 
\begin{equation}
\alpha(k)=\left\{\begin{array}{ll}\alpha_0-\frac{\displaystyle{\alpha_0-\alpha_e}}{\displaystyle{k_c}}k,&~k\le k_c \\ \alpha_e,&~k>k_c \end{array} \right. \nonumber
\end{equation}
with an initial learning rate $\alpha_0$, a final learning rate $\alpha_e$ and an epoch cap $k_c$.

In our experiments we also consider Adam (adaptive moment estimation) which is an adaptive optimisation method. It uses
estimations of first (mean) and second (uncentered variance) moments of the gradient, see \cite{kingma2017Adam} for details.
An advantage of Adam is the potential for achieving rapid training speed. While backpropagation scales the gradient uniformly
in every direction in weight space (by $\alpha$), Adam computes an individual learning rate for every weight.   

\section{Experiments and results}

For experiments on both solution approaches with different parameter variations, as well as optimisation with Adam and backpropagation,
we make use of the model problem 
\begin{equation}
\frac{d}{dx}u(x)=\lambda u(x),~~u(0)=1,
\label{TestEQ}
\end{equation}
a homogeneous first order ordinary differential equation with $\lambda\in\mathbb{R}$, $\lambda <0$. The ODE \eqref{TestEQ} has the
exact solution $u(x)=e^{\lambda x}$ and respresents a simple model for stiff phenomena involving a damping mechanism.

The numeric error $\Delta u$ shown in subsequent diagrams is defined as the $l_1$-norm of the difference between the exact solution and the
corresponding trial solution
\begin{equation}
\nonumber
\Delta u=\big\lVert u(x_i)-u_t(x_i,\vec{p})\big\rVert_1~.\nonumber
\end{equation}
With $u_t(x,\vec{p})=1+xN(x,\vec{p})$ we take the form of the trial solution for TSM
proposed in \cite{lagaris1998artificial}
to construct the cost function
\begin{align} \nonumber
E[\vec{p}]=\frac{1}{2}\bigg\lVert N(x&_i,\vec{p})+x_i\frac{\partial}{\partial x}N(x_i,\vec{p})\\ \nonumber
&-\lambda\left(1+x_iN(x_i,\vec{p})\right)\bigg\rVert_2^2~.\nonumber
\end{align}
For mTSM the trial solution $u_t(x,\vec{p})=N(x,\vec{p})$ results in the cost function
\begin{align} \nonumber
E[\vec{p}]=\frac{1}{2}&\norm{\frac{\partial}{\partial x}N(x_i,\vec{p})-\lambda N(x_i,\vec{p})}_2^2 \\ \nonumber 
+\frac{1}{2}&\big\lVert N(x_1,\vec{p})\big|_{x_1=0}-1\big\rVert_2^2~.\nonumber
\end{align}
In subsequent experiments we study $\Delta u$ with respect to several, meaningful variations of computational parameters.

The main parameters and abbreviations of the computational settings are defined as in Table \ref{parameter-names}. We use our own Fortran implementation for the neural network, the solution approaches and the optimiser, by following the proposed methods in the corresponding papers, without the use of deep learning libraries. Therefore we have total control over the computations and are able to perform investigations related to every aspect of the methods and the code.

In most subsequent experiments we used cBP instead of vBP, to reduce the amount of parameters. The learning rate for cBP is $\alpha=1$e-3 with $\beta=9$e-1.
\begin{table}[t]
\centering
\caption{\label{parameter-names}Abbreviations in the experiment section}
\vspace{-0.25cm}
\begin{tabular}{|l|c|}
\hline
Weights initialised to &   \\[-2ex]
               &  $\vec{p}^{~init}_{const}$\\[-1.25ex]
constant values &  \\
\hline
Weights initialised to &   \\[-2ex]
               &  $\vec{p}^{~init}_{rnd}$\\[-1.25ex]
random values &  \\
\hline
Numeric error for $\vec{p}^{~init}_{const}$ & $\Delta$u$_{const}$ \\
\hline
Mean value of numeric &   \\[-2ex]
               &  $\overline{\Delta u}_{rnd}$ \\[-1.25ex]
error for $\vec{p}^{init}_{rnd}$ &  \\
\hline
Trial solution method & TSM \\
\hline
modified trial solution  &   \\[-2ex]
               &  mTSM \\[-1.25ex]
method      &  \\
\hline
Backpropagation with  &   \\[-2ex]
               &  cBP \\[-1.25ex]
constant learning rate      &  \\
\hline
Backpropagation with  &   \\[-2ex]
               &  vBP \\[-1.25ex]
variable learning rate      &  \\
\hline
Adam optimisation  & Adam  \\
\hline
Number of training points & ntP \\
\hline
Number of maximal epochs & $k_{max}$ \\
\hline
Stiffness parameter & $\lambda$ \\
\hline
Left domain boundary & $x_{end}$ \\
\hline
\end{tabular}
\vspace{-0.25cm}
\end{table}
Only in optimisation comparison, vBP appears with $\alpha_0=1$e-2, $\alpha_e=1$e-3, $k_c=1$e4 and $\beta=9$e-1 as well. Adam parameters are, as employed in \cite{kingma2014Adam}, $\alpha=1$e-3, $\beta_1=9$e-1, $\beta_2=9.99$e-1 and $\epsilon=1$e-8. In addition, some experiments show averaged graphs to see the general trend with a reduced influence of fluctuations. If we do not say otherwise in the subsequent experiments, the computational parameters are fixed with one hidden layer, five hidden layer neurons, number of maximal epochs $k_{max}=1$e5, domain data $x\in[0,2]$ and stiffness parameter $\lambda=-5$. 

Concerning the following experiments, let us stress again that these are not considered to be separate or independent of each other.
We will consequently follow a line of argumentation that enables us {\em (i)} to reduce step by step the degrees of freedom in the choice of
computational settings, and {\em (ii)} to clarify the influence of individual computational parameters. In doing this we also demonstrate how
to achieve tractable results. We consider this as an important part of our work since this makes the whole approach more meaningful.

\subsection{{\bf Experiment 1}: Weight initialisation}
\label{exp1}

\begin{figure*}[!ht]
\includegraphics{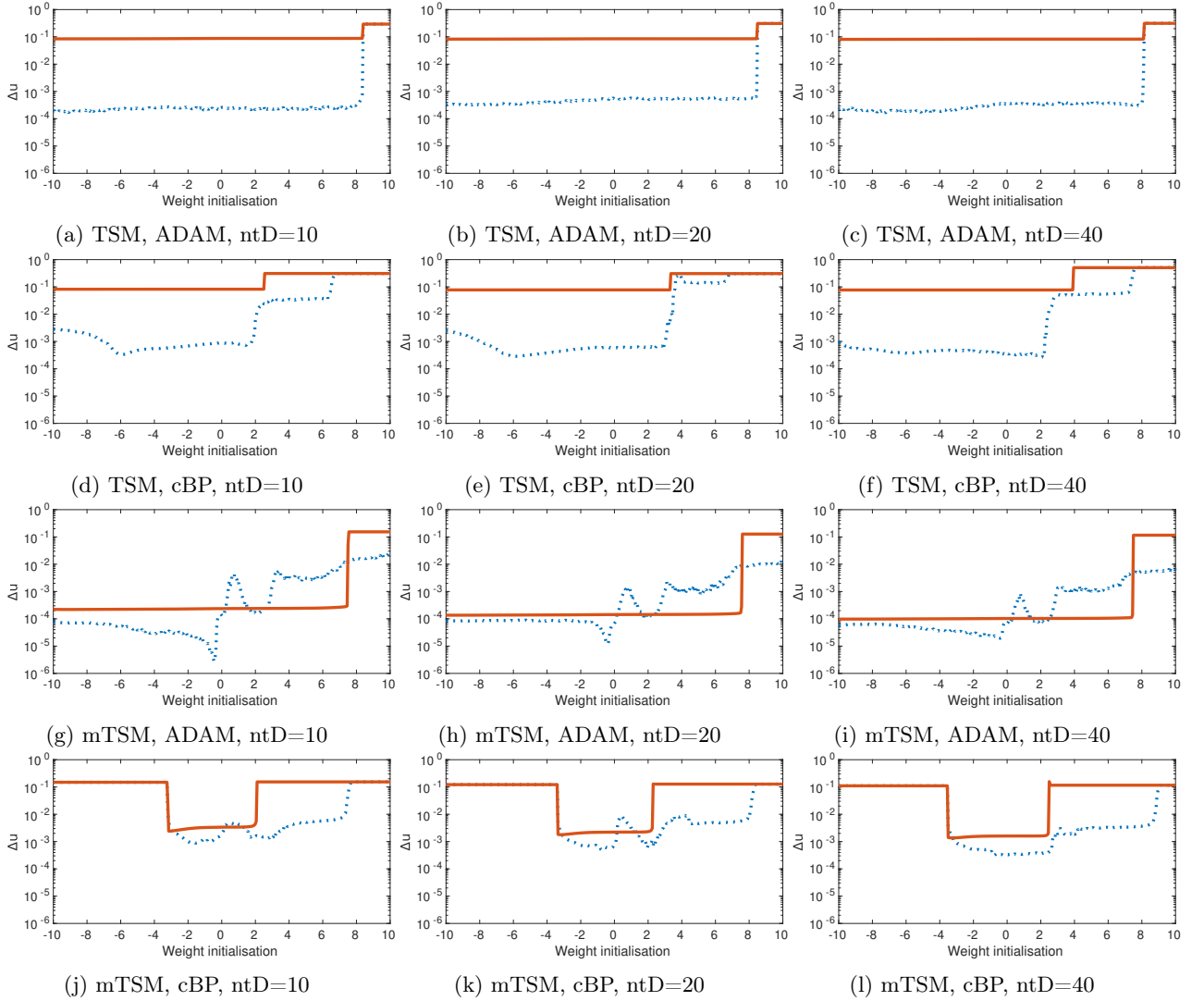}
\caption{\label{experiment1}
  {\bf Experiment \ref{exp1}}. Weight initialisation variation, (orange/solid) $\Delta u_{const}$, (blue/dotted) $\overline{\Delta u}_{rnd}$}
\end{figure*}

This experiment illustrates differences between the two weight initialisation methods, employing either $\vec{p}^{~init}_{const}$ or $\vec{p}^{~init}_{rnd}$.
We averaged 100 iterations for displaying each point in the graph depicting $\overline{\Delta u}_{rnd}$, implying that for the values given at the lower axis
we perform computations with 100 overlaid random perturbations, with random numbers in range of 1e-2, as initialisation around each point.
The averaging is important to mention, because every iteration with $\vec{p}^{~init}_{rnd}$ and exactly the same parameter setup, is expected to return different results.

Let us first comment on our choice of $\vec{p}^{~init}_{const}$. Evidently, one has to choose here some fixed value, and by
further experiments not documented here in detail, the value zero appears to be a suitable generic choice for mTSM.

Let us now consider the experiments documented in Fig.\ \ref{experiment1}.
In general, TSM with both cBP and Adam (see illustrations (a)--(f)) does not return helpful results for $\vec{p}^{~init}_{const}$ with the current parameter setup.
All experiments for TSM with $\vec{p}^{~init}_{const}$ give here uniformly a very high error (depicted by orange/solid lines), even when increasing the
number of training points.

Turning to mTSM, the overall clearly best results for $\vec{p}^{~init}_{const}$ are provided by using Adam and ntP=40 in terms of the largest stable region.
The results demonstrated in all of the experiments for mTSM with $\vec{p}^{~init}_{const}$ show that the Adam solver provides a desirable proceeding, virtually
independently of the number of training points.

When considering $\vec{p}^{~init}_{rnd}$, the Adam solver gives also for TSM reasonable results in terms of
the numerical error with a large stable region. A similar but less clear error behaviour can be observed for using Adam with mTSM.
As a general trend in all experiments with $\vec{p}^{~init}_{rnd}$, we observe that weight initialisation in a small range around zero seems to work best.

Let us also comment on illustrations (j)--(l), that we observe here the behaviour that both $\vec{p}^{~init}_{const}$ and $\vec{p}^{~init}_{rnd}$ around zero seem to work reasonably
with cBP. One may conjecture for other example ODEs, that there could be some constant initialisation and a range of random fluctuations around it
that may work well.

Since a suitable choice of $\vec{p}^{~init}_{const}$ and $\vec{p}^{~init}_{rnd}$ is important in all subsequent experiments,
we decided as a consequence of the experiments discussed here to initialise $\vec{p}^{~init}_{const}$
with zeros and $\vec{p}^{~init}_{rnd}$ with random values in range of 0 to 1e-2 from now on.

\subsection{{\bf Experiment 2}: Number of hidden layer neurons}
\label{exp2}

\begin{figure*}[!ht]
\includegraphics{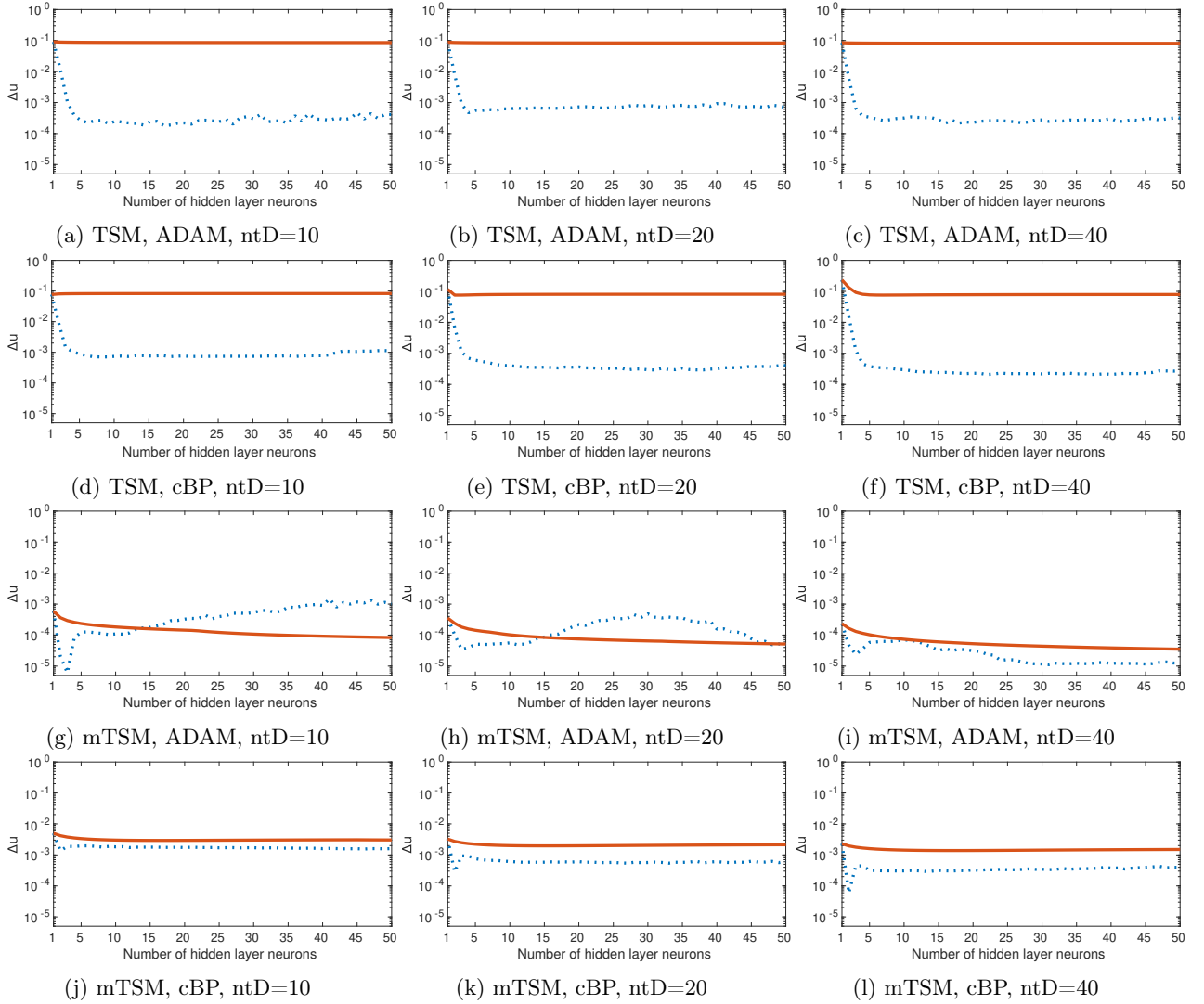}
\caption{\label{experiment2} {\bf Experiment \ref{exp2}}. Number of hidden layer neurons variation, (orange/solid) $\Delta u_{const}$, (blue/dotted) $\overline{\Delta u}_{rnd}$}
\end{figure*}
The behaviour of $\Delta u_{const}$ and $\overline{\Delta u}_{rnd}$ when increasing the number of hidden layer neurons is subject to this experiment, where $\overline{\Delta u}_{rnd}$ is averaged over 100 computations for every tested number of hidden layer neurons.

There is almost no difference between the experiments for TSM in Fig.\ \ref{experiment2} (a)--(f), they all show a similar saturating behaviour.
As discussed in the previous experiment, it is clear that we have to focus here on the random initialisation, and
for this setup we observe here desirable results for about five or more neurons.

Turning to mTSM, a higher number of hidden layer neurons leads to an increase in accuracy for Adam for larger numbers of training points explored here (ntP=40).
For smaller numbers of training points (ntP=10,20) we observe here that the number of hidden layer neurons and thus the degrees of freedom introduced
by the neural network should be in a relatively small range, e.g., about half the amount of training points.

Also for cBP the saturation value of the error is affected by increasing the amount of hidden layer neurons.
The general trend for $\overline{\Delta u}_{rnd}$ is that slightly higher accuracy is provided in this way, and that the saturation
level is visible already when using a small number of neurons.

Generally in all cases, one can clearly observe the benefit of introducing two to three or more neurons, as this leads to a significant drop
in all computed numerical errors.

As a consequence of these investigations, we employ five hidden layer neurons in the other experiments (note that this setting has also been used in
the previous experiment) as this appears to be justified by the stable solutions and the amount of computational time. 

\subsection{{\bf Experiment 3}: Number of hidden layers}
\label{exp3}

\begin{table*}[htb!]
\centering
\caption{\label{experiment3} {\bf Experiment \ref{exp3}}. Number of hidden layer variation} 
ntP=10 \par
\begin{tabular}{c|c|c|c|c|c|c|c|c}
\multirow{2}{*}{Method} & \multicolumn{4}{c|}{cBP}  & \multicolumn{4}{c}{Adam} \\
\cline{2-9}
\\[-1em]
& \multicolumn{2}{c|}{$\Delta u_{const}$}  & \multicolumn{2}{c|}{$\overline{\Delta u}_{rnd}$} & \multicolumn{2}{c|}{$\Delta u_{const}$} & \multicolumn{2}{c}{$\overline{\Delta u}_{rnd}$} \\            
\hline
Hidden Layer& TSM & mTSM & TSM & mTSM & TSM & mTSM & TSM & mTSM \\
\hline
1 & 8.25e-2 & 3.25e-3 & 7.84e-4 & 1.98e-3 & 8.74e-2 & 2.20e-4 & 2.67e-4 & 1.26e-4 \\
2 & 2.72e-3 & 2.11e-3 & 6.40e-4 & 1.15e-3 & 3.73e-3 & 5.27e-4 & 5.34e-4 & 1.84e-4 \\
3 & 6.88e-4 & 1.54e-1 & 8.15e-4 & 1.54e-1 & 2.10e-3 & 3.45e-4 & 8.69e-4 & 4.78e-4 \\
4 & 2.94e-1 & 1.54e-1 & 2.94e-1 & 1.54e-1 & 2.94e-1 & 2.93e-4 & 2.94e-1 & 1.21e-2 \\
5 & 2.94e-1 & 1.54e-1 & 2.94e-1 & 1.54e-1 & 2.94e-1 & 1.28e-1 & 2.94e-1 & 4.12e-2 \\
\end{tabular} 
\vspace{0.5cm}
\centering

ntP=20 \par
\begin{tabular}{c|c|c|c|c|c|c|c|c}
\multirow{2}{*}{Method} & \multicolumn{4}{c|}{cBP} & \multicolumn{4}{c}{Adam} \\
\cline{2-9}
\\[-1em]
& \multicolumn{2}{c|}{$\Delta u_{const}$}  & \multicolumn{2}{c|}{$\overline{\Delta u}_{rnd}$} & \multicolumn{2}{c|}{$\Delta u_{const}$} & \multicolumn{2}{c}{$\overline{\Delta u}_{rnd}$} \\       
\hline
Hidden Layer & TSM & mTSM & TSM & mTSM & TSM & mTSM & TSM & mTSM \\
\hline
1 & 7.85e-2 & 2.14e-3 & 5.43e-4 & 7.00e-4 & 8.49e-2 & 1.33e-4 & 5.51e-4 & 5.04e-5 \\
2 & 1.82e-3 & 2.24e-3 & 2.74e-4 & 5.16e-4 & 2.54e-3 & 2.06e-4 & 5.05e-4 & 2.12e-4 \\
3 & 7.76e-4 & 1.28e-1 & 3.40e-4 & 1.28e-1 & 9.91e-4 & 3.37e-4 & 9.18e-4 & 2.92e-4 \\
4 & 3.08e-1 & 1.28e-1 & 3.09e-1 & 1.28e-1 & 3.09e-1 & 1.08e-4 & 3.09e-1 & 2.26e-2 \\
5 & 3.08e-1 & 1.28e-1 & 3.09e-1 & 1.28e-1 & 3.09e-1 & 1.04e-1 & 3.09e-1 & 1.33e-2 \\
\end{tabular}
\vspace{0.5cm}
\centering

ntP=40 \par
\begin{tabular}{c|c|c|c|c|c|c|c|c}
\multirow{2}{*}{Method} & \multicolumn{4}{c|}{cBP} & \multicolumn{4}{c}{Adam} \\
\cline{2-9}
\\[-1em]
& \multicolumn{2}{c|}{$\Delta u_{const}$}  & \multicolumn{2}{c|}{$\overline{\Delta u}_{rnd}$} & \multicolumn{2}{c|}{$\Delta u_{const}$} & \multicolumn{2}{c}{$\overline{\Delta u}_{rnd}$} \\           
\hline
Hidden Layer & TSM & mTSM & TSM & mTSM & TSM & mTSM & TSM & mTSM \\
\hline
1 & 7.62e-2 & 1.54e-3 & 4.04e-4 & 3.16e-4 & 8.24e-2 & 9.35e-5 & 2.79e-4 & 6.01e-5 \\
2 & 2.33e-2 & 2.17e-3 & 2.37e-4 & 3.01e-4 & 1.55e-3 & 5.84e-5 & 4.68e-4 & 5.23e-5 \\
3 & 2.01e-3 & 1.16e-1 & 3.28e-4 & 1.16e-1 & 2.25e-3 & 1.80e-4 & 4.91e-4 & 1.19e-4 \\
4 & 3.15e-1 & 1.16e-1 & 3.15e-1 & 1.16e-1 & 3.14e-1 & 2.29e-4 & 3.13e-1 & 2.94e-2 \\
5 & 3.15e-1 & 1.16e-1 & 3.15e-1 & 1.16e-1 & 3.14e-1 & 9.46e-2 & 3.13e-1 & 1.03e-2 \\
\end{tabular}
\end{table*}
In order to focus on the impact of the number of hidden layers, we decided here to keep the number of neurons in the
hidden layers constant, employing five neurons plus an additional bias neuron in each layer.
As in previous experiments, $\overline{\Delta u}_{rnd}$ is averaged by 100 iterations.

Results in Table\ \ref{experiment3} show that one hidden layer is not always enough to provide useful results, especially for $\Delta u_{const}$ and TSM.
Increasing the number of training points (ntP) changes the number of hidden layers that give 
the best approximation in some cases, but it does not seem to have in general a highly beneficial
influence.

Turning to the most important aspect of our investigation in this experiment, one has to distinguish the effect
of increasing the number of hidden layers with respect to the individual methods mTSM and TSM. For the mTSM we find that one or two layers are sufficient to obtain -- together with Adam optimisation --
accurate and convenient results. Considering TSM our study shows a very different result, namely that each increase
in the number of hidden layers up to about three or four makes up one order of accuracy gain,
for $\vec{p}^{~init}_{const}$ and $\vec{p}^{~init}_{rnd}$.  

The latter result appears to be to some degree surprising, as the universal approximation theorem
should imply that one hidden layer could be enough to give here experimentally an accurate approximation of our solution function.
Let us recall in this context Experiment \ref{exp2}, where we have seen that an increase of the number of neurons
in one hidden layer leads to a saturation in the accuracy for $\vec{p}^{~init}_{rnd}$, while we observe here a clear improvement.
Increasing the number of neurons and using $\vec{p}^{~init}_{const}$ did not lead to reasonable results there, while $\vec{p}^{~init}_{const}$ here in combination with more hidden layers gives good results plus a significant improvement in the current study.

As a consequence of this investigation, we decided to use one hidden layer for all computations in the other
experiments, having in mind that TSM may allow an accuracy gain for more hidden layers.

\subsection{{\bf Experiment 4}: Number of epochs}
\label{exp4}

\begin{figure*}[!ht]
\includegraphics{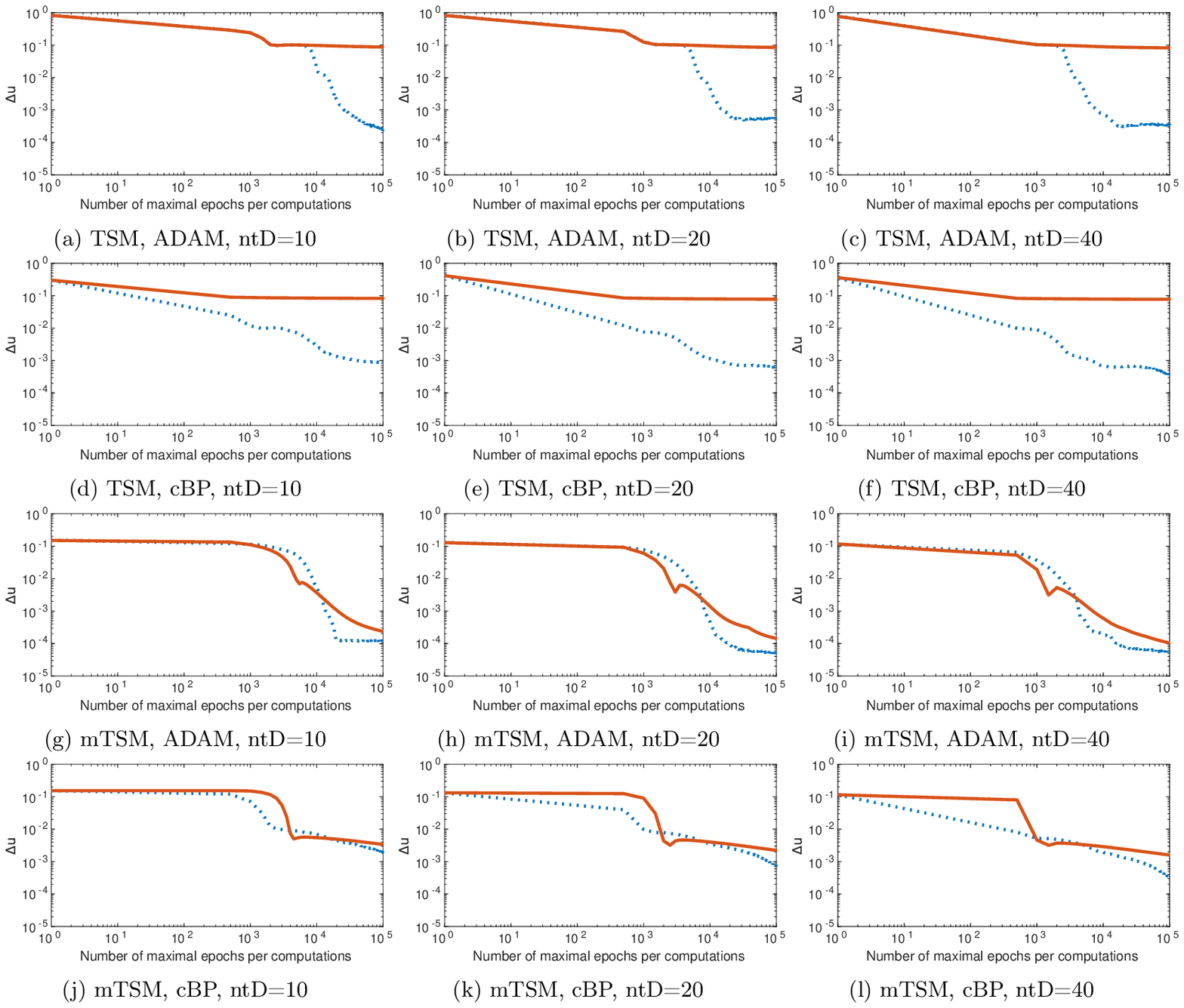}
\caption{\label{experiment4}
  {\bf Experiment \ref{exp4}}. Number of maximal epochs variation, (orange/solid) $\Delta u_{const}$, (blue/dotted) $\overline{\Delta u}_{rnd}$}
\end{figure*}
In this experiment we aim to investigate if it is possible to fix the maximal number of training epochs to a convenient
value. This relates to the question if one could bound the computational load by employing in general a small number of
training cycles. To this end, we consider the convergence of the training as a function of an increasing maximal number
of epochs $k_{max}$. In addition we illuminate the influence of the number of training points.

More precisely, we increased $k_{max}$ from 1 to 1e5 and averaged 100 iterations for one and
the same $k_{max}$. Put in other words, and to make clear the meaning of the lower axis in Fig.\ \ref{experiment4}, one
entry of the number $k_{max}$ relates to 100 corresponding complete optimisations of the neural network.
Let us note again, that in the case of $\vec{p}^{~init}_{rnd}$, the convergence behaviour can only be evaluated
by average values, and that each computation was done with a new $\vec{p}^{~init}_{rnd}$.

As can be seen in Fig.\ \ref{experiment4}, best results are returned by mTSM with Adam for both $\vec{p}^{~init}_{rnd}$
(especially ntP=20) and $\vec{p}^{~init}_{const}$ (especially ntP=40). Except for TSM and ntP=10, the Adam optimiser clearly
reaches a saturation regime showing convergence for TSM and mTSM with $\vec{p}^{~init}_{rnd}$.
For cBP, $\Delta u_{const}$ and $\overline{\Delta u}_{rnd}$, still may decrease for even higher $k_{max}$ as evaluated here.
However, let us note here that we employed in cBP a constant learning rate, for decreasing learning rates as often used
for training we may expect that a saturation regime may be observed.
However, with Adam, $\overline{\Delta u}_{rnd}$ shows a small fluctuating behaviour in the convergence regime, so that results
for non-averaged computations with $\vec{p}^{~init}_{rnd}$ may be not satisfying. The cBP optimiser together with both TSM and
mTSM shows very minor fluctuations, but also provides less good approximations. However, these tend to get better with higher
ntP.

In the context of our results, let us note that in \cite{piscopo2019solving} the authors employed 5e4 epochs. Our investigation
shows that the corresponding results are supposed to be in the convergence regime.

In conclusion, we find that $k_{max}$=1e5 as used for all other experiments is suitable to obtain useful approximations. 

\subsection{{\bf Experiment 5}: Stiffness parameter $\lambda$ (part 1) and domain size of $D$ (part 2)}
\label{exp5}

\begin{figure*}[!ht]
\includegraphics{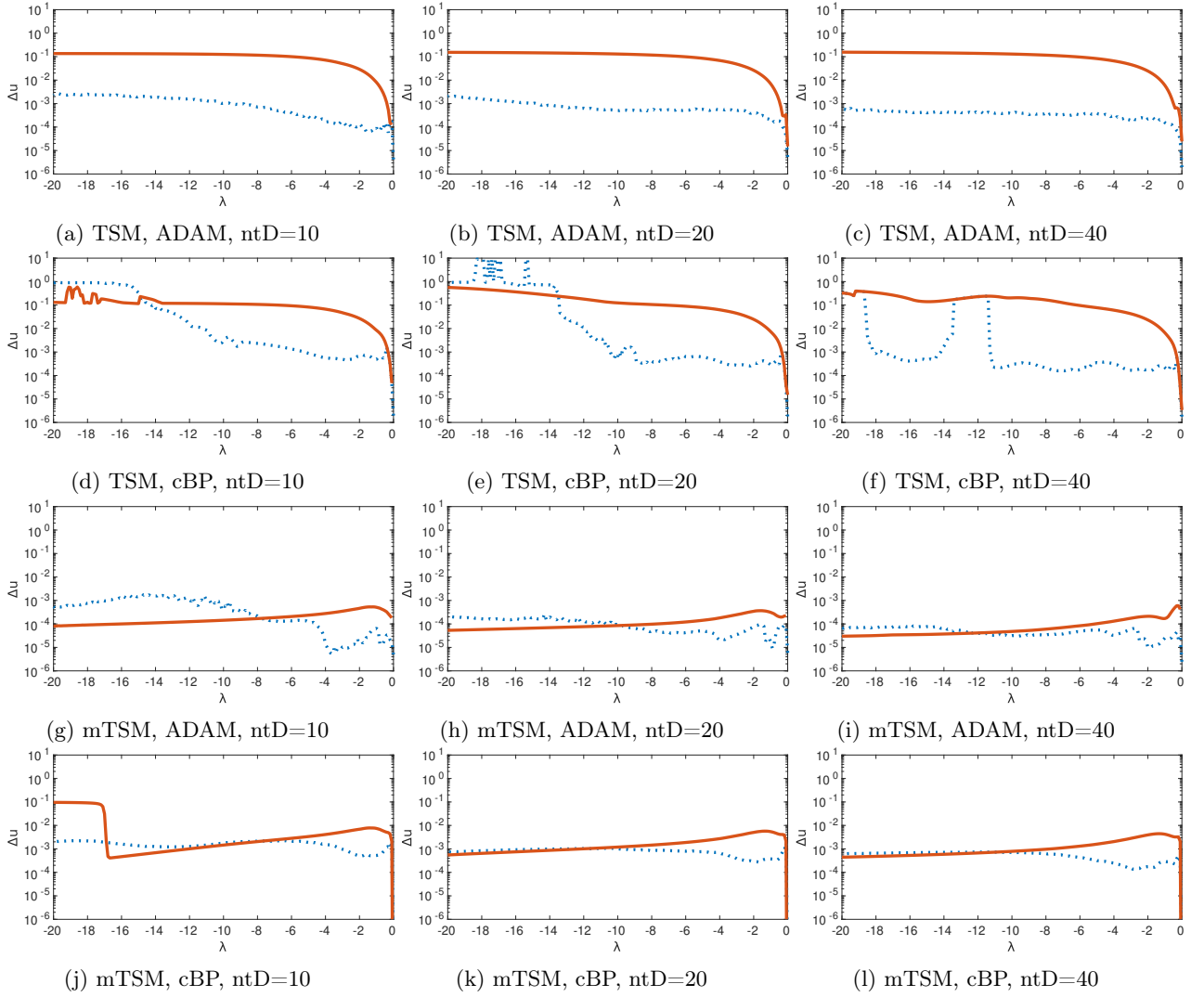}
\caption{\label{experiment5_1}{\bf Experiment \ref{exp5} (part 1)}. Stiffness parameter $\lambda$ variation, (orange/solid) $\Delta u_{const}$, (blue/dotted) $\overline{\Delta u}_{rnd}$}
\end{figure*}
\begin{figure*}[!ht]
\includegraphics{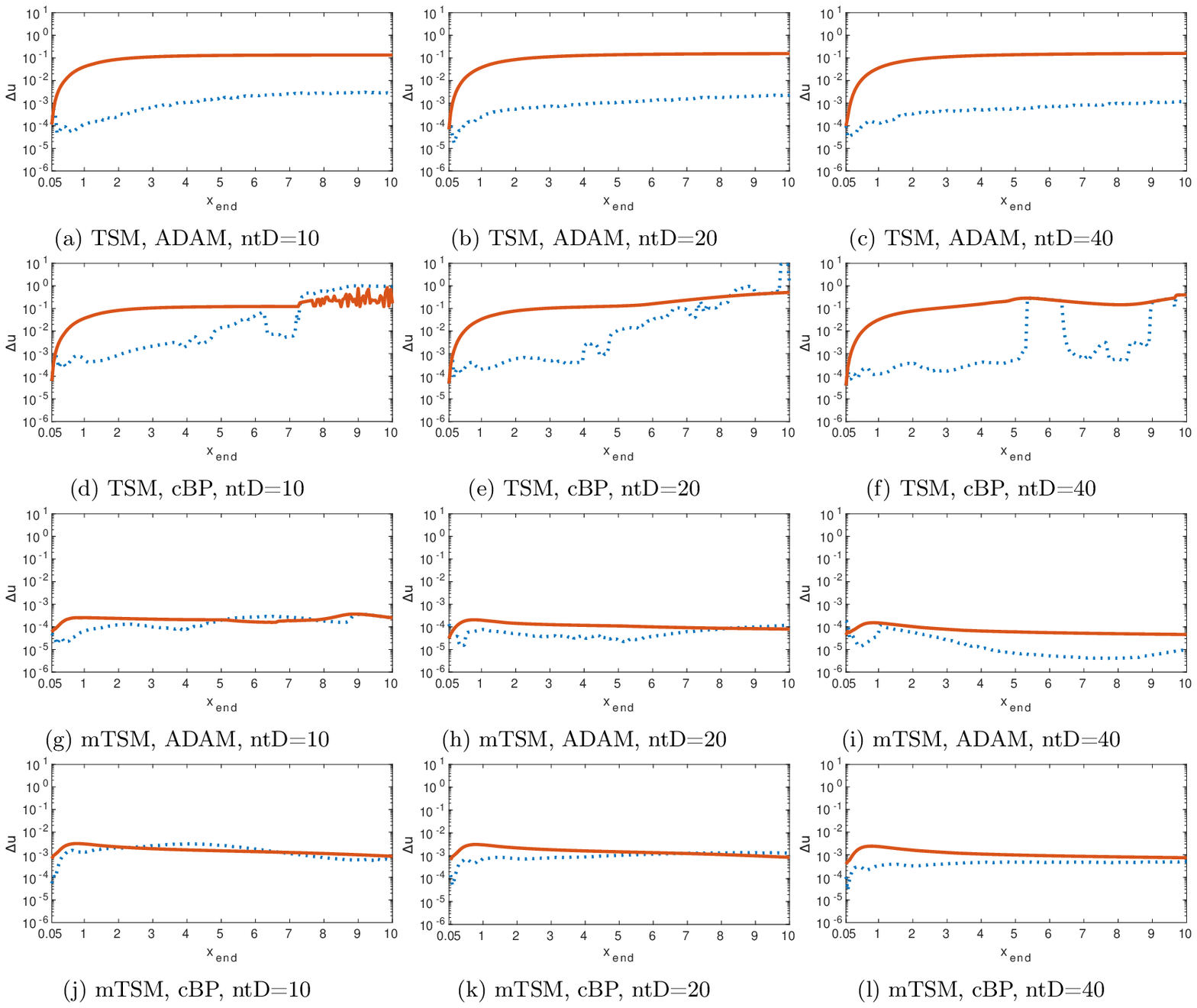}
\caption{\label{experiment5_2}{\bf Experiment \ref{exp5} (part 2)}. Domain size variation, (orange/solid) $\Delta u_{const}$, (blue/dotted) $\overline{\Delta u}_{rnd}$}
\end{figure*}
Let us now investigate the solution behaviour with respect to interesting choices of the stiffness parameter $\lambda$, and
it turns out that it makes sense to do this together with an investigation of the domain size of $D$.
Let us note that informally speaking, these parameters also impact the general trend of the exact solution in a similar way
so that it appears also from this point of view natural to evaluate them together in one experiment.

As shown in Fig.\ \ref{experiment5_2}, the influence of different domains with increasing ntP is the objective of this experiment.
Intervals used for computations are given in terms of $x\in[0,x_{end}]$, with the smallest interval being $x\in[0,5e-2]$ and then increasing
in steps of 5e-2. As also in the first experimental part here, $\overline{\Delta u}_{rnd}$ is averaged by 100 iterations for each domain.

Turning to the results, first we want to point out that for TSM, cBP and ntP=20 there are values displayed as $\overline{\Delta u}_{rnd}=9$e0,
to visualise them. In reality, these values were Not a Number (NaN), which means, that at this point at least one of the 100 averaged
iterations diverged for small values of $\lambda$ in Fig.\ \ref{experiment5_1} (c), or large domains.

Furthermore, the solution accuracy for TSM and cBP is strictly decreasing for smaller $\lambda$ and larger domains until it saturates in
unstable regions. While increasing the number of training points from ntP=10 to ntP=20 some iterations diverged, another increase to ntP=40 enlarges
the unstable region with a stabilisation in between.

In the total, we observe that there seems to be a relation between the experiments that one may roughly formulate as a relation between
$\lambda$ and domain size given by $x_{end}$ as a factor of $-2$. We also conjecture, that the higher the values of $-\lambda$ and $x_{end}$,
the more neurons or layers are required for a convenient solution.
As a consequence of these experiments, we decided to fix $\lambda=-5$ and $x\in[0,2]$ for all computations in the other experiments.

\subsection{{\bf Experiment 6}: Optimisation methods}
\label{exp6}

\begin{figure*}[!ht]
\includegraphics{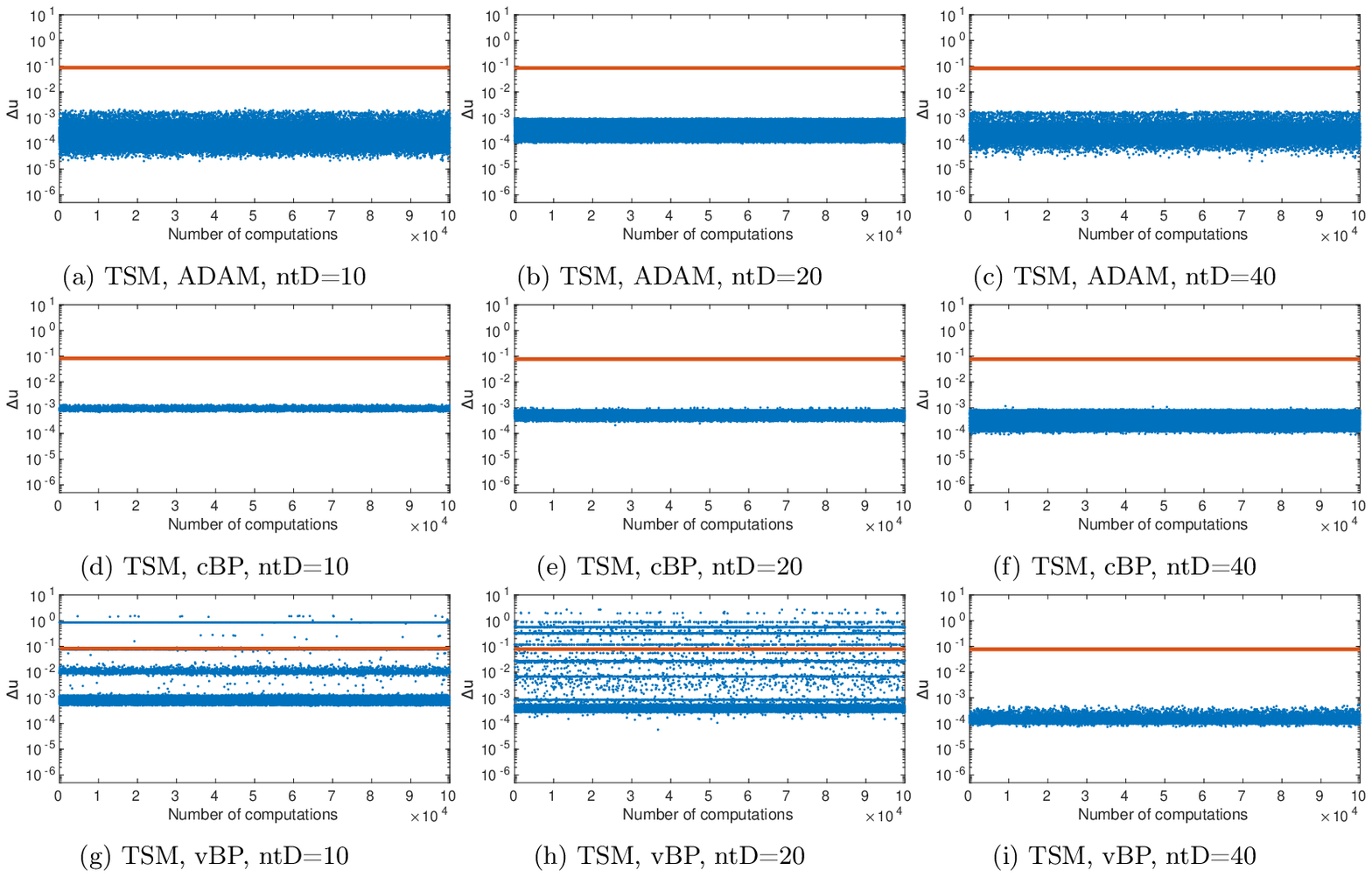}
\caption{\label{experiment6_1}{\bf Experiment \ref{exp6}}. Optimiser comparison (part 1), (orange/solid) $\Delta u_{const}$, (blue/dotted) $\Delta u_{rnd}$}
\end{figure*}
\begin{figure*}[!ht]
\includegraphics{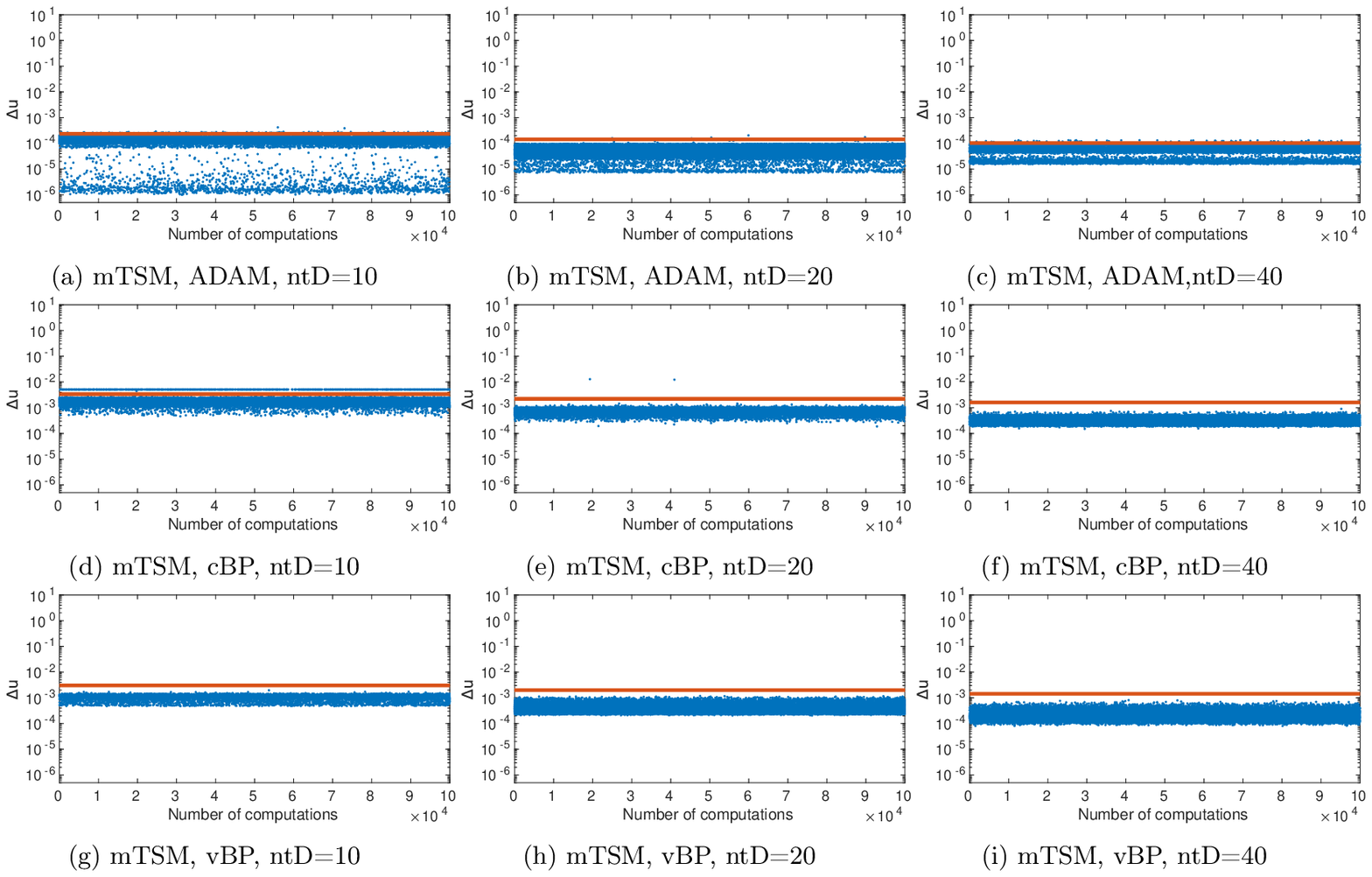}
\caption{\label{experiment6_2}{\bf Experiment \ref{exp6}}. Optimiser comparison (part 2), (orange/solid) $\Delta u_{const}$, (blue/dotted) $\Delta u_{rnd}$}
\end{figure*}

\begin{table*}
\centering
\caption{\label{experiment6_3}{\bf Experiment \ref{exp6}}. Optimiser comparison (part 3), quantitative data for $\Delta u_{rnd}$}
\begin{tabular}{|c|l|l|c|c|c|c|c|}
\hline
\multicolumn{1}{|l|}{method} & \multicolumn{1}{c|}{\begin{tabular}[c]{@{}c@{}}training\\ data\end{tabular}} & optimiser & \begin{tabular}[c]{@{}c@{}}mean\\ value\end{tabular} & \begin{tabular}[c]{@{}c@{}}standard\\ deviation\end{tabular} & \begin{tabular}[c]{@{}c@{}}10\%-\\ quantile\end{tabular} & \begin{tabular}[c]{@{}c@{}}20\%-\\ quantile\end{tabular} & \begin{tabular}[c]{@{}c@{}}30\%-\\ quantile\end{tabular} \\ \hline
\multirow{9}{*}{TSM} & \multirow{3}{*}{ntP=10} & Adam & 2.70e-4 & 1.89e-4 & 9.54e-5 & 1.31e-4 & 1.66e-4 \\ \cline{3-8} 
 &  & cBP & 8.79e-4 & 5.56e-5 & 8.16e-4 & 8.40e-4 & 8.54e-4 \\ \cline{3-8} 
 &  & vBP & 8.89e-2 & 2.32e-1 & 6.32e-4 & 7.09e-4 & 8.05e-4 \\ \cline{2-8} 
 & \multirow{3}{*}{ntP=20} & Adam & 5.49e-4 & 2.38e-4 & 1.88e-4 & 2.80e-4 & 3.91e-4 \\ \cline{3-8} 
 &  & cBP & 6.07e-4 & 1.11e-4 & 4.48e-4 & 5.08e-4 & 5.55e-4 \\ \cline{3-8} 
 &  & vBP & 4.78e-2 & 1.44e-1 & 4.04e-4 & 4.64e-4 & 4.70e-4 \\ \cline{2-8} 
 & \multirow{3}{*}{ntP=40} & Adam & 3.55e-4 & 1.86e-4 & 1.40e-4 & 1.96e-4 & 2.49e-4 \\ \cline{3-8} 
 &  & cBP & 3.60e-4 & 1.69e-4 & 1.64e-4 & 2.06e-4 & 2.46e-4 \\ \cline{3-8} 
 &  & vBP & 1.54e-4 & 3.31e-5 & 1.10e-4 & 1.22e-4 & 1.34e-4 \\ \hline
\multirow{9}{*}{mTSM} & \multirow{3}{*}{ntP=10} & Adam & 1.24e-4 & 2.51e-5 & 1.06e-4 & 1.10e-4 & 1.13e-4 \\ \cline{3-8} 
 &  & cBP & 2.01e-3 & 4.45e-4 & 1.64e-3 & 1.77e-3 & 1.89e-3 \\ \cline{3-8} 
 &  & vBP & 1.24e-3 & 1.04e-4 & 1.18e-3 & 1.23e-3 & 1.25e-3 \\ \cline{2-8} 
 & \multirow{3}{*}{ntP=20} & Adam & 5.01e-5 & 1.51e-5 & 3.35e-5 & 3.75e-5 & 4.11e-5 \\ \cline{3-8} 
 &  & cBP & 7.46e-4 & 1.67e-4 & 5.73e-4 & 6.02e-4 & 6.24e-4 \\ \cline{3-8} 
 &  & vBP & 4.02e-4 & 1.11e-4 & 2.72e-4 & 3.10e-4 & 3.40e-4 \\ \cline{2-8} 
 & \multirow{3}{*}{ntP=40} & Adam & 5.67e-5 & 1.30e-5 & 4.76e-5 & 5.03e-5 & 5.23e-5 \\ \cline{3-8} 
 &  & cBP & 3.22e-4 & 7.26e-5 & 2.53e-4 & 2.70e-4 & 2.79e-4 \\ \cline{3-8} 
 &  & vBP & 2.29e-4 & 6.28e-5 & 1.70e-4 & 1.83e-4 & 1.93e-4 \\ \hline
\end{tabular}
\end{table*}

The final experiment in this paper compares Adam, cBP and vBP optimisation for TSM and mTSM, depending on ntP=10,20,40 with the other computational parameters fixed to one hidden layer, five hidden layer neurons, $k_{max}=1$e5, $\lambda=-5$ and $x\in[0,2]$.
Fig.\ \ref{experiment6_1} and \ref{experiment6_2} show 1e5 (non-averaged) computed results for each parameter setup and weight initialisation.  \par

Previous experiments led to the conclusion, that TSM in combination with $\vec{p}^{~init}_{const}$ only provides unstable solutions for the
chosen parameter setup. Therefore, when evaluating TSM, we will only refer to the non-averaged numeric error $\Delta u_{rnd}$ for $\vec{p}^{~init}_{rnd}$. \par

To start the evaluation with TSM and Adam, there are almost no visible differences between ntP=10 and ntP=40, with a large difference between
the best and the least good approximation, see first row in Fig.\ \ref{experiment6_1}. Only for ntP=20 the solutions tend to be more similar. \par

In contrast, the difference between the best and the least good approximation for TSM and cBP grows by one order of magnitude with a higher number of training points
while simultaneously the accuracy for the best approximations increases, cf.\ second row in the figure. \par

The reason we show results on vBP only in this final experiment (see third row in the figure) is, that the efficiency of an adaptive step size method may be in general highly
dependent on the used step size model and parameters. However, the results turn out to be interesting. In combination with ntP=10, vBP and TSM reveal several minima far away
from the best approximation. Even more minima appear for a training points increase to ntP=20. However, another increase to ntP=40 stabilises the solutions.
In addition, ntP=40 provides the best approximations for TSM and vBP. One may conjecture here, that either one has here to reach a critical number of training points, or
that the weight initialisation here is not adequate together with lower ntP.\par

Now we turn to mTSM and Adam, see first row in Fig.\ \ref{experiment6_2}. We find $\vec{p}^{~init}_{const}$ to show useful results ($\Delta u_{const}$)
and a small gain in accuracy for higher ntP. For $\vec{p}^{~init}_{rnd}$, we find the best approximations throughout the whole experiment to be provided by ntP=10.
However, most of the 1e5 computed results appear around a less good (but still reasonable) accuracy with only a few results peaking further in accuracy.
Increasing the number of training points to ntP=20 and ntP=40 results in a drop of accuracy from the former best solutions, while overall the results become
more similar. \par

For mTSM and cBP, see second row in the figure, we find a similar behaviour of $\Delta u_{const}$, similarly to the case mTSM and Adam. The solutions become
slightly more accurate and similar with higher ntP. However both weight initialisation methods can not compete with the combination mTSM and Adam. \par

Now for mTSM and vBP as displayed by the last row in the figure, we find stable results for all ntP, which is in sharp contrast to TSM and vBP.
Again, $\Delta$u$_{const}$ behaves like the other computations for mTSM, and we find similarities in the overall behaviour of $\Delta u_{rnd}$ compared to TSM and cBP.
Increasing ntP leads to slightly better approximations, while the difference between the best and the least good approximation grows. \par

Turning to Table \ref{experiment6_3}, we now discuss the stochastic quantities for $\vec{p}^{~init}_{rnd}$, related to the results shown in Fig.\ \ref{experiment6_1} and \ref{experiment6_2}. We focus on the results for random weight initialisation as the diagrams have shown the constant weight initialisation to always return the same numerical error. The 1e5 complete computations (optimisations) should sufficiently support the meaning of the analysed data. \par 
Regarding the mean value, Adam has the overall smallest value and seems to be the best choice. However, for TSM and ntP=40, cBP almost equals Adam with vBP outperforming Adam in this specific setting. That result is particular interesting, since vBP shows for TSM and both ntP=10 and ntP=20 very limited approximations. In contrast to TSM, Adam dominates for mTSM the lowest mean value without any exception. \par
The former statement however does only hold partially when it comes to the standard deviation. Here, TSM seems to favour cBP over Adam, again with vBP for ntP=40 to pass downwards. Excluding vBP for ntP=10 and ntP=20, the standard deviation in the other cases are in an acceptable range. That is, the mean value and standard deviation could be suitable when evaluating stability and reliability. However, it can be difficult to specify the term reliability. The lower the numerical error, the better the approximation. None\-the\-less, defining a threshold needs justification and discussion on how the neural network methods behave compared to standard numerical algorithms like Runge-Kutta. \par
We also take different quantiles (10\%,20\%,30\%) into account. The percentage specifies the relative amount of data points which appear below the quantile value itself. Although several minima for TSM and vBP in Fig.\ \ref{experiment5_1} (g),(h) appear to be less useful than the lowest one, all quantiles for these cases are better than for the same settings with cBP. The situation for mTSM is the same, while Adam outperforms both cBP and vBP in this context. Therefore one may find that further adjusting the optimisation parameters for vBP can in general lead to perform better than cBP. However, it is questionable if this would also perform better than Adam. We find all quantiles values to be good in case of Adam optimisation. In this sense, we consider Adam here as the most reliable optimiser.

Concluding, the overall best performance related to the numeric error shows mTSM and Adam for both $\vec{p}^{~init}_{const}$ and $\vec{p}^{~init}_{rnd}$.
Although TSM and vBP appear to have some stability flaws for lower ntP, it stabilises for ntP=40. Overall, both vBP and cBP can not compete with Adam and mTSM.

\section{{\bf Conclusion and future work}}

When solving the stiff model ODE with feedforward neural networks, the solution reliability depends on a variety of parameters. We find the weight
initialisation to have a major influence. While the initialisation with zeros does not provide reasonable appromixations for TSM with one hidden layer, it is capable
to work reasonably well for mTSM. First setting the weights to small random values shows the best results with Adam and mTSM, although the use of more training points
may yield less suitable results. This may indicate an overfitting and could be resolved by employing more neurons or other adjustments.
This may be a subject for a future study.

However, our work also indicates that all the investigated issues may have to be considered together as a complete package, i.e., the investigated aspects may
not be evaluated completely independent of each other. Even after a detailed investigation as provided here it seems not to be possible to single out an individual aspect
that dominates the overall accuracy and reliability.

We tend to favour the combination of Adam and mTSM in further computationally oriented research, since it provides the best approximations for both weight initialisation
methods. Future research may also include theoretical work, e.g., on sensitivity and different trial solution forms for TSM. One main goal in this context is to decrease the
variation of possible solutions together with an increase of the solution accuracy. 

Moreover, our third experiment has shown that it may make sense to investigate deep networks,
since these could result in a significant accuracy gain, reminding of higher order effects in
classic numerical analysis. 

Furthermore, our future work will include more difficult differential equations with a focus on initial value problems and the improvement of constant weight initialisation.

\section*{Acknowledgement}
This publication was funded by the Graduate Research School (GRS) of the Brandenburg University of Technology Cottbus-Senftenberg. This work is part of the
Research Cluster Cognitive Dependable Cyber Physical Systems.

\end{document}